\documentclass[final]{IEEEtran} 
\IEEEoverridecommandlockouts
\usepackage{cite}
\usepackage{amsmath,amssymb,amsfonts}
\usepackage{algorithmic}
\usepackage{caption}
\usepackage{subcaption}
\usepackage{graphicx}
\usepackage{textcomp}
\usepackage{xcolor}
\usepackage{array}
\usepackage{hyperref}
\usepackage{etoolbox}
\appto\UrlBreaks{\do\-}
\hypersetup{draft} % https://www.overleaf.com/learn/latex/Questions/What_does_%22%5Cpdfendlink_ended_up_in_different_nesting_level_than_%5Cpdfstartlink%22_mean%3F

\def\BibTeX{{\rm B\kern-.05em{\sc i\kern-.025em b}\kern-.08em
    T\kern-.1667em\lower.7ex\hbox{E}\kern-.125emX}}
\begin{document}

\title{Lincoln AI Computing Survey (LAICS) Update\\
\thanks{This material is based upon work supported by the Assistant Secretary of Defense for Research and Engineering under Air Force Contract No. FA8702-15-D-0001. Any opinions, findings, conclusions or recommendations expressed in this material are those of the author(s) and do not necessarily reflect the views of the Assistant Secretary of Defense for Research and Engineering.}
}

\author{\IEEEauthorblockN{Albert Reuther, Peter Michaleas, Michael Jones, \\Vijay Gadepally, Siddharth Samsi, and Jeremy Kepner} \\
\IEEEauthorblockA{\textit{MIT Lincoln Laboratory Supercomputing Center} \\
Lexington, MA, USA \\
\{reuther,pmichaleas,michael.jones,vijayg,sid,kepner\}@ll.mit.edu}
}

\maketitle

\begin{abstract}

This paper is an update of the survey of AI accelerators and processors from past four years, which is now called the Lincoln AI Computing Survey -- LAICS (pronounced ``lace''). As in past years, this paper collects and summarizes the current commercial accelerators that have been publicly announced with peak performance and peak power consumption numbers. The performance and power values are plotted on a scatter graph, and a number of dimensions and observations from the trends on this plot are again discussed and analyzed. Market segments are highlighted on the scatter plot, and zoomed plots of each segment are also included. Finally, a brief description of each of the new accelerators that have been added in the survey this year is included. 

\end{abstract}

\begin{IEEEkeywords}
Machine learning, GPU, TPU, tensor, dataflow, CGRA, accelerator, embedded inference, computational performance
\end{IEEEkeywords}

\section{Introduction}

A number of announcements, releases, and deployments of artificial intelligence (AI) accelerators from startups and established technology companies have occurred in the past year. Perhaps most notable is the emergence of very large foundation models that are able to generate prose, poetry, images, etc. based on training using vast amounts of data usually collected via internet data crawls. Much technical press has been focused on how effective the resulting tools will be for various tasks, but also there is much discussion about the training of these models. But from an accelerator perspective, it is the very same accelerators that are aimed towards training more modestly sized models that are used for training these very large models. The very large models are just using many more accelerators simultaneously in a synchronous parallel manner, and they are interconnected with very high bandwidth networks. But beyond that news, not much has changed in the overall trends and landscape. Hence, this paper just updates what was discussed in last year's survey. 

For much of the background of this study, please refer to one of the previous IEEE-HPEC papers that our team has published~\cite{reuther2022ai,reuther2021ai,reuther2020survey,reuther2019survey}. This background includes an explanation of the AI ecosystem architecture, the history of the emergence of AI accelerators and accelerators in general, a more detailed explanation of the survey scatter plots, and a discussion of broader observations and trends.

%\begin{landscape}
\begin{figure*}[!bth]
    \centering
    \includegraphics[width=\textwidth]{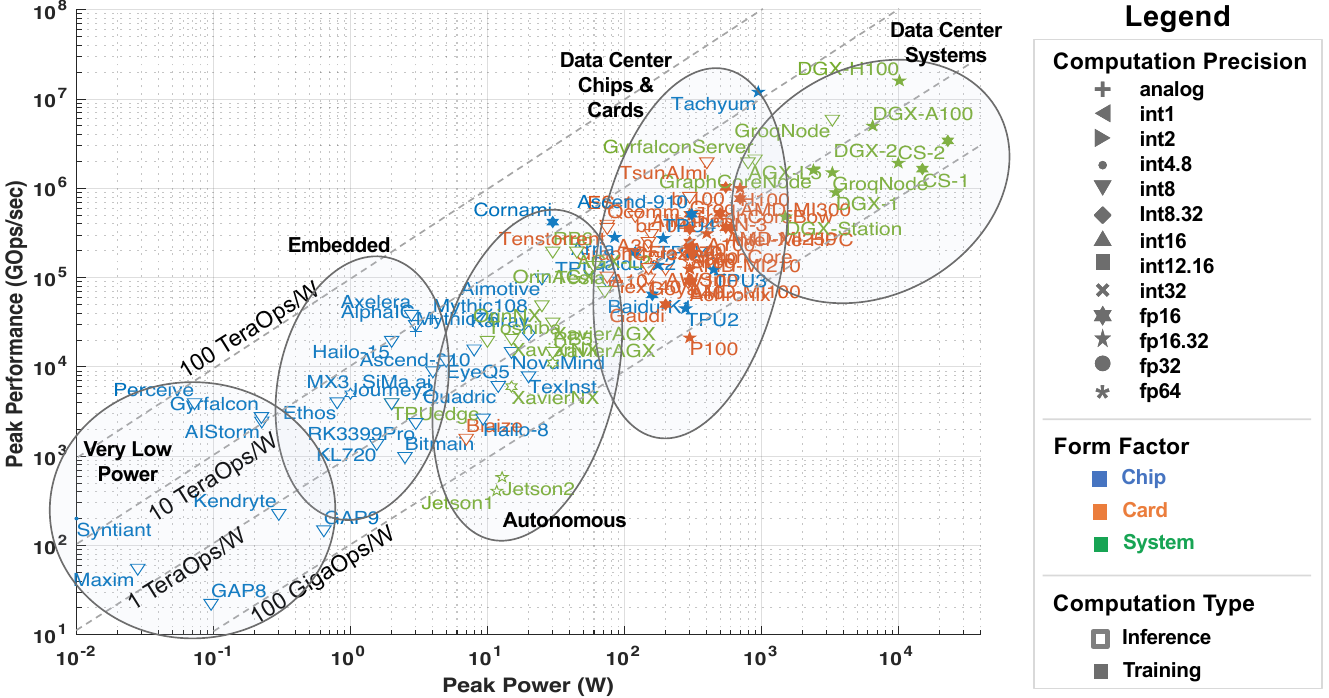}
    \caption{Peak performance vs. power scatter plot of publicly announced AI accelerators and processors.}
    \label{fig:PeakPerformancePower}
  \end{figure*}
%\end{landscape}

%\begin{landscape}
\begin{figure*}[!thb]
    \begin{subfigure}{.48\textwidth} %
	\centering
        \includegraphics[width=0.98\textwidth]{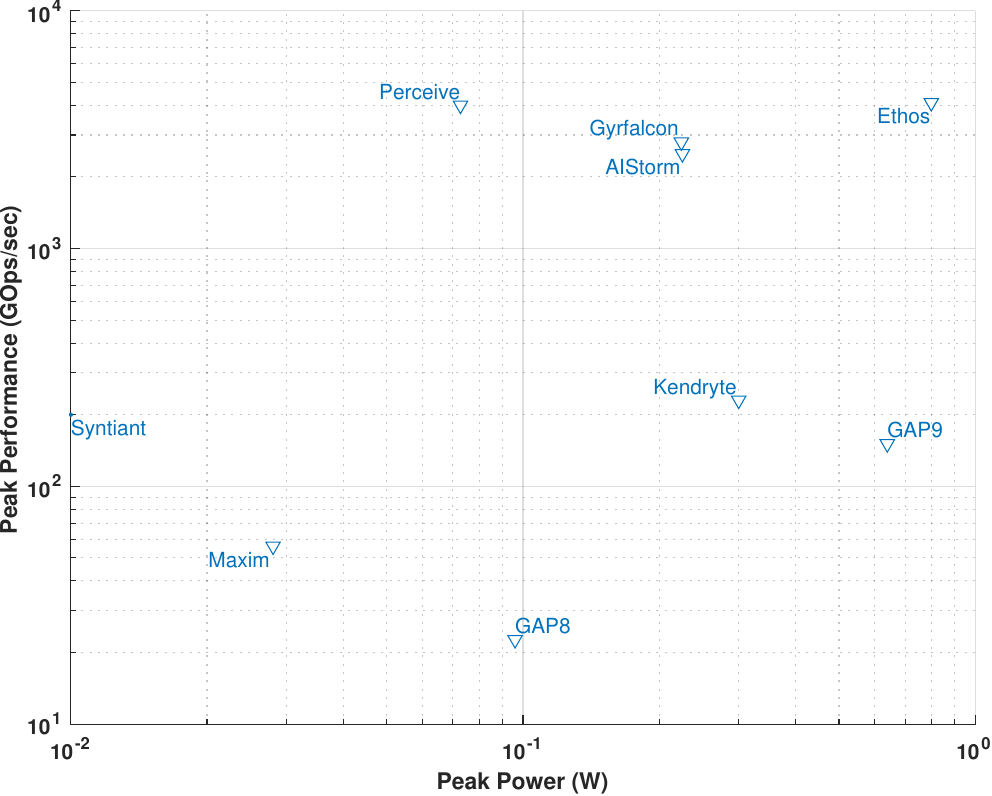}
		\caption{}
    \end{subfigure}
    \begin{subfigure}{.48\textwidth} %%
	\centering
        \includegraphics[width=0.98\textwidth]{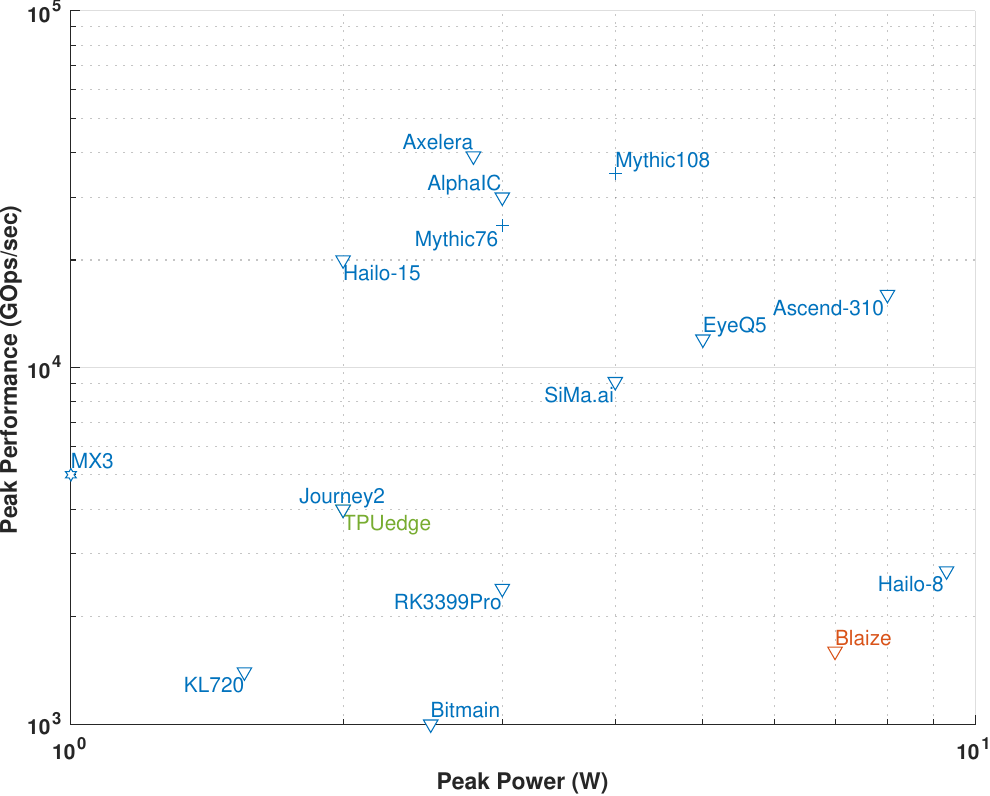}
		\caption{}
    \end{subfigure}
%\newline
    
    \begin{subfigure}{.48\textwidth} %
	\centering
        \includegraphics[width=0.98\textwidth]{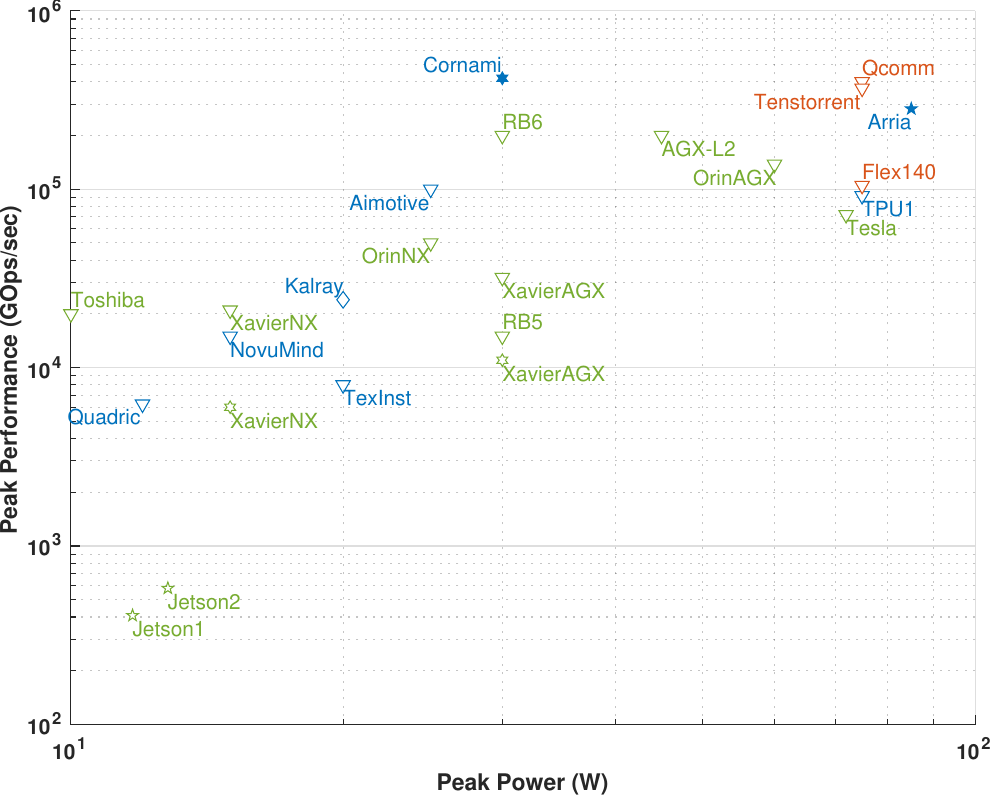}
		\caption{}
    \end{subfigure}
    \begin{subfigure}{.48\textwidth} %
	\centering
        \includegraphics[width=0.98\textwidth]{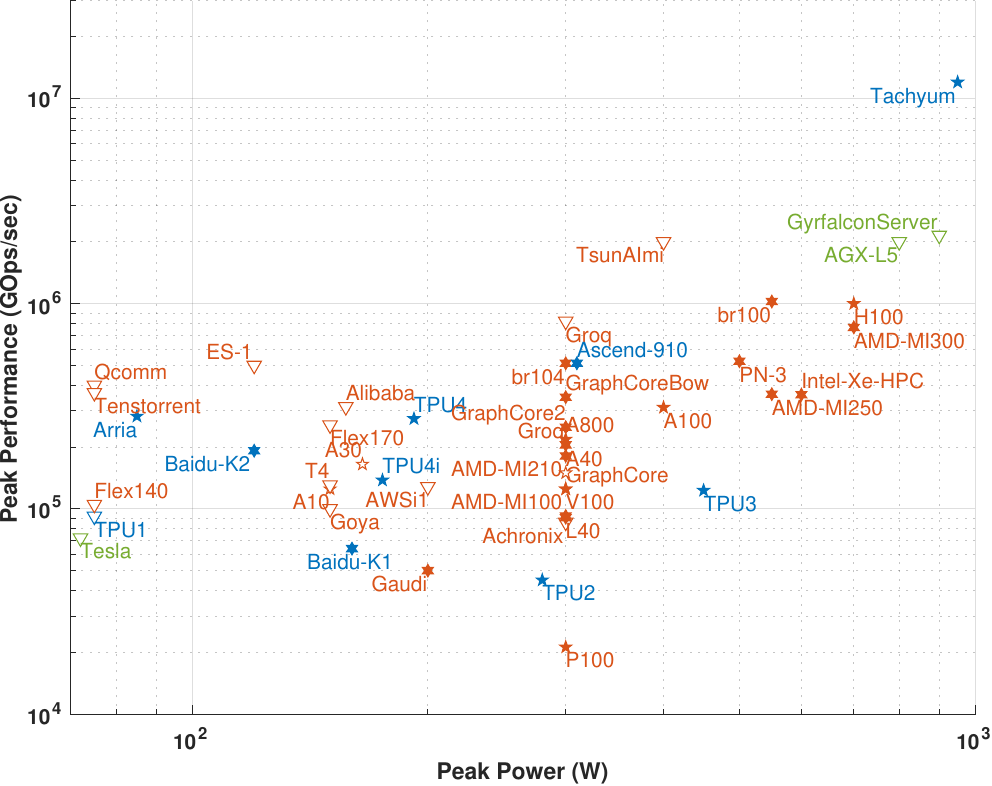}
		\caption{}
    \end{subfigure}
%\newline

	\centering
    \begin{subfigure}{.48\textwidth} %
        \includegraphics[width=0.98\textwidth]{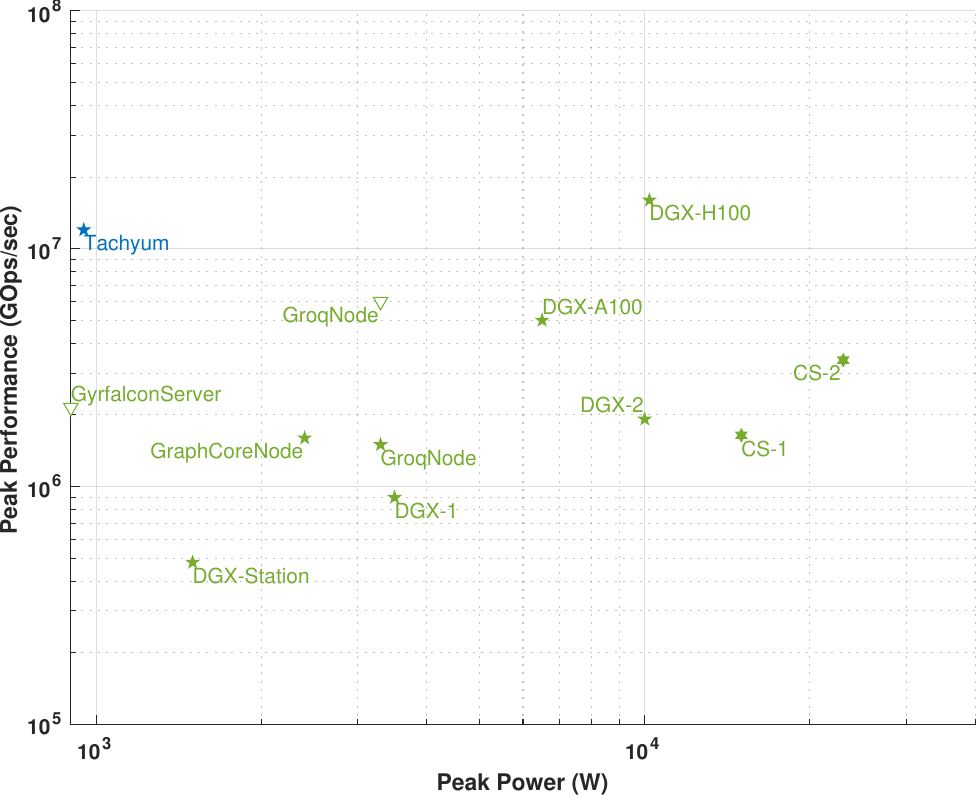}
		\caption{}
    \end{subfigure}

    \caption{Zoomed regions of peak performance vs. peak power scatter plot: \textbf{(a)} very low power, \textbf{(b)} embedded, 
    \textbf{(c)} autonomous, \textbf{(d)} data center chips and cards, \textbf{(e)} data center systems.}
    \label{fig:PeakPerformancePowerZoomed}
\end{figure*}
%\end{landscape}

\section{Survey of Processors}

This paper is an update to IEEE-HPEC papers from the past four years~\cite{reuther2022ai,reuther2021ai,reuther2020survey,reuther2019survey}. 
This survey continues to cast a wide net to include accelerators and processors for a variety of applications including defense and national security AI/ML edge applications. 
The survey collects information on all of the numerical precision types that an accelerator supports, but for most of them, their best inference performance is in int8 or fp16/bf16, so that is what usually is plotted. This survey gathers performance and power information from publicly available materials including research papers, technical trade press, company benchmarks, etc. The key metrics of this public data are plotted in Figure~\ref{fig:PeakPerformancePower}, which graphs recent processor capabilities (as of Summer 2023) mapping peak performance vs. power consumption, and Table~\ref{tab:acceleratorlist} summarizes some of the important metadata of the accelerators, cards, and systems, including the labels used in Figure~\ref{fig:PeakPerformancePower}. 

The x-axis indicates peak power, and the y-axis indicate peak giga-operations per second (GOps/s), both on a logarithmic scale. The computational precision of the processing capability is depicted by the geometric marker used. The form factor is depicted by  color, which shows the package for which peak power is reported. Blue corresponds to a single chip; orange corresponds to a card; and green corresponds to entire systems (single node desktop and server systems). Finally, the hollow geometric objects are peak performance for inference-only accelerators, while the solid geometric figures are performance for accelerators that are designed to perform both training and inference. 

A reasonable categorization of accelerators follows their intended application, and the five categories are shown as ellipses on the graph, which roughly correspond to performance and power consumption: Very Low Power for wake word detection, speech processing, very small sensors, etc.; Embedded for cameras, small UAVs and robots, etc.; Autonomous for driver assist services, autonomous driving, and autonomous robots; Data Center Chips and Cards; and Data Center Systems. A zoomed in scatter plot for each of these categories is shown in the subfigures of Figure~\ref{fig:PeakPerformancePowerZoomed}.

\begingroup
\setlength{\tabcolsep}{2pt} % Default value: 6pt

\begin{table}
  \centering
  %\scriptsize
  \tiny
  \caption{List of accelerator metadata and labels for plots.}
  \label{tab:acceleratorlist}

\begin{tabular}{| l | l | c | c | c | c |} \hline 

 \textbf{Company} & \textbf{Product} & \textbf{Label} & \textbf{Technology} & \textbf{Form Factor} & \textbf{References} \\ \hline 
     Achronix & VectorPath S7t-VG6 & Achronix & FPGA & Card & \cite{roos2019fpga}  \\ \hline  
     Aimotive & aiWare3 & Aimotive & dataflow & Chip & \cite{aimotive2018aiware3}  \\ \hline  
     AIStorm & AIStorm & AIStorm & dataflow & Chip & \cite{merritt2019aistorm}  \\ \hline  
     Alibaba & HanGuang 800 & Alibaba & dataflow & Card & \cite{peng2019alibaba}  \\ \hline  
     AlphaIC & RAP-E & AlphaIC & dataflow & Chip & \cite{clarke2018indo}  \\ \hline  
     Amazon & Inferentia & AWSi1 & dataflow & Card & \cite{hamilton2018aws,cloud2020deep}  \\ \hline  
     AMD & MI100 & AMD-MI100 & GPU & Card & \cite{smith2021amd}  \\ \hline  
     AMD & MI210 & AMD-MI210 & GPU & Card & \cite{smith2021amd}  \\ \hline  
     AMD & MI250 & AMD-MI250 & GPU & Card & \cite{smith2021amd}  \\ \hline  
     AMD & MI300 & AMD-MI300 & GPU & Card & \cite{morgan2023amd,morgan2023third}  \\ \hline  
     ARM & Ethos N77 & Ethos & dataflow & Chip & \cite{schor2020arm}  \\ \hline  
     Axelera & Axelera Test Core & Axelera & dataflow & Chip & \cite{ward2022axelera}  \\ \hline  
     Baidu & Baidu Kunlun 200 & Baidu-K1 & dataflow & Chip & \cite{ouyang2021kunlun,merritt2018baidu,duckett2018baidu}  \\ \hline  
     Baidu & Baidu Kunlun II & Baidu-K2 & dataflow & Chip & \cite{shilov2021baidu}  \\ \hline  
     Biren Technology & br100 & br100 & GPU & Card & \cite{peckham2022chinese,shilov2023chinese}  \\ \hline  
     Biren Technology & br104 & br104 & GPU & Card & \cite{peckham2022chinese,shilov2023chinese}  \\ \hline  
     Bitmain & BM1880 & Bitmain & dataflow & Chip & \cite{wheeler2019bitmain}  \\ \hline  
     Blaize & El Cano & Blaize & dataflow & Card & \cite{demler2020blaize}  \\ \hline  
     Canaan & Kendrite K210 & Kendryte & CPU & Chip & \cite{gwennap2019kendryte}  \\ \hline  
     Cerebras & CS-1 & CS-1 & dataflow & System & \cite{hock2019introducing}  \\ \hline  
     Cerebras & CS-2 & CS-2 & dataflow & System & \cite{trader2021cerebras}  \\ \hline  
     Cornami & Cornami & Cornami & dataflow & Chip & \cite{cornami2020cornami}  \\ \hline  
     Enflame & Cloudblazer T10 & Enflame & CPU & Card & \cite{clarke2019globalfoundries}  \\ \hline  
     Esperanto & ET-SoC-1 & ES-1 & CPU & Card & \cite{ditzel2022accelerating,schor2021look}  \\ \hline  
     Google & TPU Edge & TPUedge & tensor & System & \cite{tpu2019edge}  \\ \hline  
     Google & TPU1 & TPU1 & tensor & Chip & \cite{jouppi2020domain,teich2018tearing}  \\ \hline  
     Google & TPU2 & TPU2 & tensor & Chip & \cite{jouppi2020domain,teich2018tearing}  \\ \hline  
     Google & TPU3 & TPU3 & tensor & Chip & \cite{jouppi2021ten,jouppi2020domain,teich2018tearing}  \\ \hline  
     Google & TPU4i & TPU4i & tensor & Chip & \cite{jouppi2021ten}  \\ \hline  
     Google & TPU4 & TPU4 & tensor & Chip & \cite{peckham2022google}  \\ \hline  
     GraphCore & C2 & GraphCore & dataflow & Card & \cite{gwennap2020groq,lacey2017preliminary}  \\ \hline  
     GraphCore & C2 & GraphCoreNode & dataflow & System & \cite{graphcore2020dell}  \\ \hline  
     GraphCore & Colossus Mk2 & GraphCore2 & dataflow & Card & \cite{ward2020graphcore}  \\ \hline  
     GraphCore & Bow-2000 & GraphCoreBow & dataflow & Card & \cite{tyson2022graphcore}  \\ \hline  
     GreenWaves & GAP8 & GAP8 & dataflow & Chip & \cite{greenwaves2020gap,turley2020gap9}  \\ \hline  
     GreenWaves & GAP9 & GAP9 & dataflow & Chip & \cite{greenwaves2020gap,turley2020gap9}  \\ \hline  
     Groq & Groq Node & GroqNode & dataflow & System & \cite{hemsoth2020groq}  \\ \hline  
     Groq & Groq Node & GroqNode & dataflow & System & \cite{hemsoth2020groq}  \\ \hline  
     Groq & Tensor Streaming Processor & Groq & dataflow & Card & \cite{gwennap2020groq,abts2020think}  \\ \hline  
     Groq & Tensor Streaming Processor & Groq & dataflow & Card & \cite{gwennap2020groq,abts2020think}  \\ \hline  
     Gyrfalcon & Gyrfalcon & Gyrfalcon & PIM & Chip & \cite{ward2019gyrfalcon}  \\ \hline  
     Gyrfalcon & Gyrfalcon & GyrfalconServer & PIM & System & \cite{hpcwire2020solidrun}  \\ \hline  
     Habana & Gaudi & Gaudi & dataflow & Card & \cite{gwennap2019habanagaudi,medina2020habana}  \\ \hline  
     Habana & Goya HL-1000 & Goya & dataflow & Card & \cite{gwennap2019habanagoya,medina2020habana,gwennap2019habanagoya}  \\ \hline  
     Hailo & Hailo-8 & Hailo-8 & dataflow & Chip & \cite{ward2019details}  \\ \hline  
     Hailo & Hailo-15H & Hailo-15 & dataflow & Chip & \cite{ward2023hailo}  \\ \hline  
     Horizon Robotics & Journey2 & Journey2 & dataflow & Chip & \cite{horizon2020journey}  \\ \hline  
     Huawei HiSilicon & Ascend 310 & Ascend-310 & dataflow & Chip & \cite{huawei2020ascend310}  \\ \hline  
     Huawei HiSilicon & Ascend 910 & Ascend-910 & dataflow & Chip & \cite{huawei2020ascend910}  \\ \hline  
     Intel & Arria 10 1150 & Arria & FPGA & Chip & \cite{abdelfattah2018dla,hemsoth2018intel}  \\ \hline  
     Intel & Mobileye EyeQ5 & EyeQ5 & dataflow & Chip & \cite{demler2020blaize}  \\ \hline  
     Intel & Xe-HPC & Intel-Xe-HPC & GPU & Card & \cite{shilov2021intels,intel2021intel,shilov2022intels}  \\ \hline  
     Intel & Flex140 & Flex140 & GPU & Card & \cite{morgan2022different}  \\ \hline  
     Intel & Flex170 & Flex170 & GPU & Card & \cite{morgan2022different}  \\ \hline  
     Kalray & Coolidge & Kalray & manycore & Chip & \cite{dupont2019kalray, clarke2020nxp}  \\ \hline  
     Kneron & KL720 & KL720 & dataflow & Chip & \cite{ward2021kneron}  \\ \hline  
     Maxim & Max 78000 & Maxim & dataflow & Chip & \cite{ward2020maxim,jani2021maxim,clay2022benchmarking}  \\ \hline  
     MemryX & MX3 & MX3 & dataflow & Chip & \cite{leibson2023adding,vicinanza2022startup}  \\ \hline  
     Mythic & M1076 & Mythic76 & PIM & Chip & \cite{ward2021mythic,hemsoth2018mythic,fick2018mythic}  \\ \hline  
     Mythic & M1108 & Mythic108 & PIM & Chip & \cite{ward2021mythic,hemsoth2018mythic,fick2018mythic}  \\ \hline  
     NovuMind & NovuTensor & NovuMind & dataflow & Chip & \cite{freund2019novumind,yoshida2018novumind}  \\ \hline  
     NVIDIA & Ampere A10 & A10 & GPU & Card & \cite{morgan2021nvidia}  \\ \hline  
     NVIDIA & Ampere A100 & A100 & GPU & Card & \cite{krashinsky2020nvidia}  \\ \hline  
     NVIDIA & Ampere A800 & A800 & GPU & Card & \cite{shilov2023nvidias}  \\ \hline  
     NVIDIA & Ampere A30 & A30 & GPU & Card & \cite{morgan2021nvidia}  \\ \hline  
     NVIDIA & Ampere A40 & A40 & GPU & Card & \cite{morgan2021nvidia}  \\ \hline  
     NVIDIA & DGX Station & DGX-Station & GPU & System & \cite{alcorn2017nvidia}  \\ \hline  
     NVIDIA & DGX-1 & DGX-1 & GPU & System & \cite{alcorn2017nvidia,cutress2018nvidias}  \\ \hline  
     NVIDIA & DGX-2 & DGX-2 & GPU & System & \cite{cutress2018nvidias}  \\ \hline  
     NVIDIA & DGX-A100 & DGX-A100 & GPU & System & \cite{campa2020defining}  \\ \hline  
     NVIDIA & DGX-H100 & DGX-H100 & GPU & System & \cite{mujtaba2022nvidia}  \\ \hline  
     NVIDIA & H100 & H100 & GPU & Card & \cite{smith2022nvidia}  \\ \hline  
     NVIDIA & Jetson AGX Xavier & XavierAGX & GPU & System & \cite{smith2019nvidia}  \\ \hline  
     NVIDIA & Jetson NX Orin & OrinNX & GPU & System & \cite{funk2022nvidia,nvidia2022embedded}  \\ \hline  
     NVIDIA & Jetson AGX Orin & OrinAGX & GPU & System & \cite{funk2022nvidia,nvidia2022embedded}  \\ \hline  
     NVIDIA & Jetson TX1 & Jetson1 & GPU & System & \cite{franklin2017nvidia}  \\ \hline  
     NVIDIA & Jetson TX2 & Jetson2 & GPU & System & \cite{franklin2017nvidia}  \\ \hline  
     NVIDIA & Jetson Xavier NX & XavierNX & GPU & System & \cite{smith2019nvidia}  \\ \hline  
     NVIDIA & DRIVE AGX L2 & AGX-L2 & GPU & System & \cite{hill2020nvidia}  \\ \hline  
     NVIDIA & DRIVE AGX L5 & AGX-L5 & GPU & System & \cite{hill2020nvidia}  \\ \hline  
     NVIDIA & L40 & L40 & GPU & Card & \cite{techpowerup2023nvidia}  \\ \hline  
     NVIDIA & Pascal P100 & P100 & GPU & Card & \cite{pascal2018nvidia,smith201816gb}  \\ \hline  
     NVIDIA & T4 & T4 & GPU & Card & \cite{kilgariff2018nvidia}  \\ \hline  
     NVIDIA & Volta V100 & V100 & GPU & Card & \cite{volta2019nvidia,smith201816gb}  \\ \hline  
     Perceive & Ergo & Perceive & dataflow & Chip & \cite{mcgregor2020perceive}  \\ \hline  
     Preferred Networks & MN-3 & PN-3 & multicore & Card & \cite{preferred2020mncore, cutress2019preferred}  \\ \hline  
     Quadric & q1-64 & Quadric & dataflow & Chip & \cite{firu2019quadric}  \\ \hline  
     Qualcomm & Cloud AI 100 & Qcomm & dataflow & Card & \cite{ward2020qualcomm,mcgrath2019qualcomm}  \\ \hline  
     Qualcomm & QRB5165 & RB5 & GPU & System & \cite{crowe2020qualcomm}  \\ \hline  
     Qualcomm & QRB5165N & RB6 & GPU & System & \cite{qualcomm2023robotics}  \\ \hline  
     Rockchip & RK3399Pro & RK3399Pro & dataflow & Chip & \cite{rockchip2018rockchip}  \\ \hline  
     SiMa.ai & SiMa.ai & SiMa.ai & dataflow & Chip & \cite{gwennap2020machine}  \\ \hline  
     Syntiant & NDP101 & Syntiant & PIM & Chip & \cite{mcgrath2018tech,merritt2018syntiant}  \\ \hline  
     Tachyum & Prodigy & Tachyum & CPU & Chip & \cite{shilov2022tachyum}  \\ \hline  
     Tenstorrent & Tenstorrent & Tenstorrent & multicore & Card & \cite{gwennap2020tenstorrent}  \\ \hline  
     Tesla & Tesla FSC & Tesla & dataflow & System & \cite{talpes2020compute,wikichip2020fsd}  \\ \hline  
     Texas Instruments & TDA4VM & TexInst & dataflow & Chip & \cite{ward2020ti,ti2021tda4vm,demler2020ti}  \\ \hline  
     Toshiba & 2015 & Toshiba & multicore & System & \cite{merritt2019samsung}  \\ \hline  
     Untether & TsunAImi & TsunAImi & PIM & Card & \cite{gwennap2020untether}  \\ \hline  

\end{tabular}
\end{table}
\endgroup

For most of the accelerators, their descriptions and commentaries have not changed since last year so please refer to the papers of the last four years for descriptions and commentaries. Several new releases are included in this update.

\begin{itemize} 

\item Based on similar technology of its Cloud AI 100 accelerator, Qualcomm has released two versions of its robotics AI system platform, the RB5 and RB6, in the past few years. Both are competing in the same low power system-on-a-chip market as the NVIDIA Jetson product line, and are aimed at integration in applications including robotics, driver assist, modest UAVs, etc.~\cite{crowe2020qualcomm,qualcomm2023robotics}.

\item The Memryx MX3 AI accelerator chip is a startup that was spun out of the University of Michigan. It is designed to be deployed with a host CPU to greatly speed up AI inference, consuming about 1W of power. It computes activations with bf16 numerical precision, and store model parameter weight at 4-bit, 8-bit, and 16-bit integer precisions, which can be set on a layer-by-layer basis~\cite{leibson2023adding,vicinanza2022startup}.

\item On the heels of it's Hailo-8 AI accelerator, Hailo has released a lower power variant, the Hailo-15. The Hailo-15 targets the Internet Protocol (IP) camera market, and it is a SoC that includes a CPU, a digital signal processor (DSP) accelerator, and a neural accelerator, which all draw less than 2W~\cite{ward2023hailo}.

\item Startup Esperanto Technologies has released their first processor accelerator called the ET-SoC-1. Each chip is comprised of 1,088 64-bit ET-Minion RISC-V cores, each of which have scalar, vector, and tensor units along with L1 cache/scratchpad memory. Their key application is training and inference for recommender systems, which have a balanced mix of scalar, vector, and tensor operations ~\cite{ditzel2022accelerating,schor2021look}.

\item Baidu has started deploying its second-generation Kunlun accelerator, Kunlun II. Baidu touted that the Kunlun II is 2-3 times faster than the original Kunlun~\cite{shilov2021baidu}.

\item The Chinese GPU startup Biren emerged from stealth mode to announce and release two high performance GPUs: the BR100 and BR104. The BR104 is a single die GPU, while the BR100 combines two dies/chiplets in the same package~\cite{peckham2022chinese,shilov2023chinese}.

\item AMD has announced the followup to their Instinct MI250 GPU called the Instinct MI300A, which will be a multi-chiplet CPU-GPU Accelerated Processing Unit (APU) integrated package. The announcement showed package photos of two CPU dies integrated with six GPU dies.~\cite{morgan2023amd,morgan2023third}

\item While Intel announced their high-end AI GPU a few years ago, details continued to be scarce until this past year. Enough performance numbers were announced for the Intel Xe-HPC (codename Ponte Vecchio) to include it in this year's survey~\cite{shilov2021intels,intel2021intel,shilov2022intels}. Along with the Xe-HPC, Intel also announced and started shipping two inference-oriented GPU cards, the Flex 140 and Flex 170~\cite{morgan2022different}. 

\item After announcing and shipping their Hopper H100 GPUs in systems at the end of 2022, NVIDIA has started shipping DGX servers, which integrate eight H100 GPUs~\cite{mujtaba2022nvidia}. NVIDIA has also released a high-performance Ampere GPU, the A800, that is aimed at the Chinese market, which reportedly performs at approximately 70\% peak performance of the A100~\cite{shilov2023nvidias}. Finally, NVIDIA has released a new Ada Lovelace GPU family which is aimed at the data center inference and graphics rendering (gaming) farm markets. The first specifications were released for the L40 GPU, which are included in this survey~\cite{techpowerup2023nvidia}. 

\end{itemize}

\section{Summary}

This paper updates the Lincoln AI Computing Survey (LAICS) of deep neural network accelerators that span from extremely low power through embedded and autonomous applications to data center class accelerators for inference and training. We presented the new full scatter plot along with zoomed in scatter plots for each of the major deployment/market segments, and we discussed some new additions for the year. The rate of announcements and releases has continued to be consistent as companies compete for various embedded, data center, cloud, and on-premises HPC deployments.

\section{Data Availability}

The data spreadsheets and references that have been collected for this study and its papers will be posted at \url{https://github.com/areuther/ai-accelerators} after they have cleared the release review process. 

\section*{Acknowledgement}

We express our gratitude to 
Masahiro Arakawa, Bill Arcand, Bill Bergeron, David Bestor, Bob Bond, Chansup Byun, Nathan Frey, Vitaliy Gleyzer, Jeff Gottschalk, Michael Houle, Matthew Hubbell, Hayden Jananthan, Anna Klein, David Martinez, Joseph McDonald, Lauren Milechin, Sanjeev Mohindra, Paul Monticciolo, Julie Mullen, Andrew Prout, Stephan Rejto, Antonio Rosa, Charles Yee, and Marc Zissman
for their support of this work. 


\bibliographystyle{IEEEtran} 
\bibliography{AIAcceleratorTrends2023}

%%\end{thebibliography}

\end{document}